\documentclass[conference]{IEEEtran}
\IEEEoverridecommandlockouts

\usepackage{cite}
\usepackage{amsmath,amssymb,amsfonts}
\usepackage{algorithmic}
\usepackage{graphicx}
\usepackage{textcomp}
\usepackage{xcolor}
\usepackage{url}
\usepackage{booktabs}
\usepackage{subfigure}

\def\BibTeX{{\rm B\kern-.05em{\sc i\kern-.025em b}\kern-.08em
    T\kern-.1667em\lower.7ex\hbox{E}\kern-.125emX}}
    
\begin{document}

\title{Drone Stereo Vision for Radiata Pine Branch Detection and Distance Measurement: Integrating SGBM and Segmentation Models\\}

\author{\IEEEauthorblockN{Yida Lin, Bing Xue, Mengjie Zhang}
\IEEEauthorblockA{
\small
\textit{Centre for Data Science and Artificial Intelligence} \\
\textit{Victoria University of Wellington, Wellington, New Zealand}\\
linyida\texttt{@}myvuw.ac.nz, 
bing.xue\texttt{@}vuw.ac.nz, 
mengjie.zhang\texttt{@}vuw.ac.nz}
\and
\IEEEauthorblockN{Sam Schofield, Richard Green}
\IEEEauthorblockA{
\small
\textit{Department of Computer Science and Software Engineering} \\
\textit{Canterbury University, Canterbury, New Zealand}\\
sam.schofield\texttt{@}canterbury.ac.nz, 
richard.green\texttt{@}canterbury.ac.nz}
}

\IEEEpubidadjcol

\maketitle

\begin{abstract}
Manual pruning of radiata pine trees presents significant safety risks due to their substantial height and the challenging terrains in which they thrive. To address these risks, this research proposes the development of a drone-based pruning system equipped with specialized pruning tools and a stereo vision camera, enabling precise detection and trimming of branches. Deep learning algorithms, including YOLO and Mask R-CNN, are employed to ensure accurate branch detection, while the Semi-Global Matching algorithm is integrated to provide reliable distance estimation. The synergy between these techniques facilitates the precise identification of branch locations and enables efficient, targeted pruning. Experimental results demonstrate that the combined implementation of YOLO and SGBM enables the drone to accurately detect branches and measure their distances from the drone. This research not only improves the safety and efficiency of pruning operations but also makes a significant contribution to the advancement of drone technology in the automation of agricultural and forestry practices, laying a foundational framework for further innovations in environmental management.
\end{abstract}

\begin{IEEEkeywords}
Tree Pruning with Drone, Semi-Global Matching, Supervised Learning, Stereo Vision.
\end{IEEEkeywords}

\section{Introduction}
Pinus radiata, commonly known as radiata pine, is a highly valuable species extensively cultivated in New Zealand due to its rapid growth and versatile applications in forestry and timber industries. This species is essential for producing high-quality timber used in construction, paper manufacturing, and other wood-based products, significantly contributing to the economy\cite{van2013national}\cite{mason2023impacts}. For instance, in New South Wales, Australia, the radiata pine industry contributed approximately \$3 billion to the economy in 2021-2022, highlighting its economic importance\footnote{https://www.dpi.nsw.gov.au/dpi/climate/climate-vulnerability-assessment/forestry/radiata-pine}. However, to ensure the trees grow with strong, straight trunks and produce clear wood, which doesn't have knots, regular pruning is necessary. Traditionally performed manually, tree pruning and trimming are hazardous occupations globally, posing significant challenges and dangers. According to Tree Care Industry Magazine\footnote{https://tcimag.tcia.org/safety/tree-work-safety-by-the-numbers/}, in the United States alone, the Bureau of Labor Statistics reports a fatality rate of 110 per 100,000 tree trimmers and pruners, which is about 30 times higher than the average across all industries. Moreover, non-fatal injury rates for tree workers are also substantially higher, at approximately 239 injuries per 10,000 workers, compared to 89 per 10,000 across all industries. It is also challenging to find people who want to do the work because it is hard and dangerous.

To effectively replace manual labor in branch pruning, we aim to develop a fully autonomous drone system. Existing drone pruning systems typically require manual operation and are limited to cutting only thicker branches. Additionally, they often rely on expensive auxiliary equipment such as LiDAR sensors, which significantly hinders their widespread adoption \cite{molina2017aerial}. To overcome these limitations, we propose the development of a drone equipped with a stereo camera and a pruning tool capable of automatically detecting and pruning branches as thin as 10mm in diameter. This system utilizes the stereo camera for both branch identification and distance measurement, enabling fully autonomous pruning operations. By streamlining the design and enhancing sensor technology, our approach aims to make drone-based pruning more accessible and cost-effective, thereby improving precision and efficiency in forestry management. This advancement in autonomous drone technology not only enhances forestry practices but also offers a versatile and economical solution for various applications beyond forestry, eliminating the need for extensive manual control or expensive auxiliary equipment.

\begin{figure}[htbp]
\centering
\includegraphics[width=8cm]{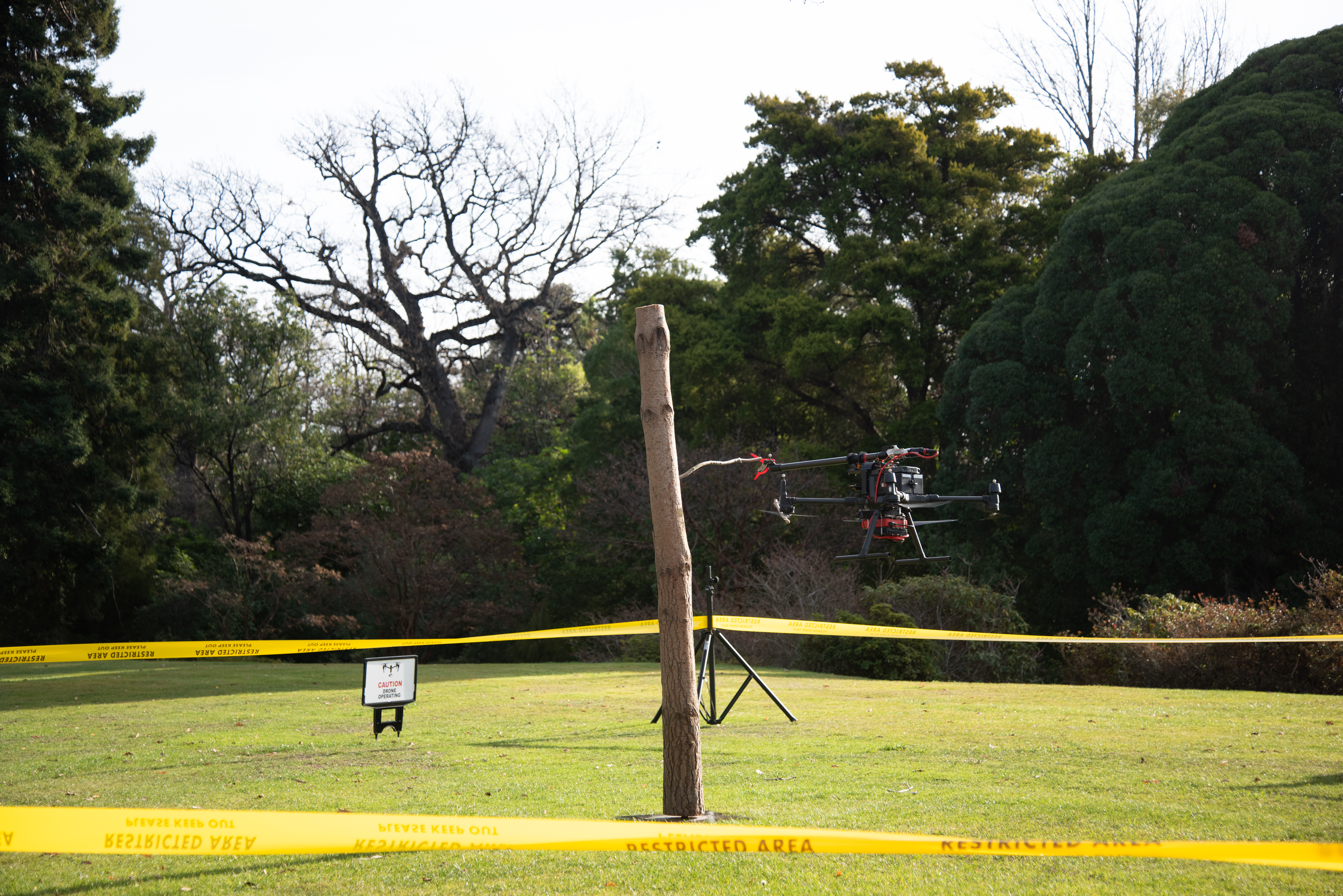}
\caption{The drone, equipped with a ZED mini camera for stereo vision and a pruning tool autonomously detects and prunes branches of radiata pine. The ZED mini camera enables the drone to accurately identify the branches, while the pruning tool  precisely prunes them.}
\label{drone}
\end{figure}

To ensure this research is comprehensive and self-contained, it is imperative to include detailed information on all relevant aspects, particularly the structural components of the drones utilized in this research. For a comprehensive overview of these components, please refer to the detailed information on drones available at \url{https://ucvision.org.nz/drones/}. This link provides critical insights into the design and specifications of the drones, with their testing process depicted in Fig. \ref{drone}.

The big research project is a collaborative effort involving multiple institutions, each contributing specific expertise. While other institutions focus on the physical construction of the drones, our research is concentrated on developing the vision detection and measurement algorithms for the cameras mounted on these drones. These algorithms are pivotal for accurately detecting branches and determining their precise positions in three-dimensional space, which is essential for guiding the drone’s pruning tool to prune branches effectively. The primary objective of our research is to enhance the accuracy and reliability of branch detection, thereby improving the overall efficiency and safety of the autonomous pruning process.

\section{Related Work}

We propose employing a computer vision system to simultaneously perform branch detection and distance measurement tasks, offering a more efficient and cost-effective alternative \cite{zou2023object}. Therefore, our research focus will concentrate on a standalone computer vision approach. Building on these advancements, we are committed to developing a novel method for real-time branch detection and depth estimation, enabling drones to accurately determine the spatial positions of tree branches. By leveraging advanced stereo vision techniques optimized for computational efficiency, this method aims to provide a scalable, cost-effective solution that enhances pruning precision while reducing the equipment load on drones \cite{bazargani2024automation}.

Implementing this approach necessitates addressing several key components: accurate object detection and image segmentation to identify tree branches, generation of depth maps to estimate spatial positions, and the establishment of robust performance metrics to evaluate system efficiency and accuracy. In the following subsections, we delve into these components in detail, reviewing existing methodologies and elucidating how our proposed solutions contribute to and advance the current state of the field.

\subsection{Object detection and image segmentation}
Object detection\cite{zou2023object} and image segmentation\cite{minaee2021image} are both critical tasks in computer vision area. Object detection primarily focuses on identifying and locating objects within an image, typically by marking their positions with bounding boxes. Segmentation, on the other hand, takes this a step further by dividing the image into distinct regions, accurately delineating the shape and boundaries of objects. 

In this research, the focus extends beyond merely identifying the positional information of tree branches to include the acquisition of detailed locational data of surrounding points. This requirement necessitates a transition from conventional object detection methods to more precise image segmentation techniques \cite{sharma2022survey}. By utilizing segmentation on drone-captured imagery, this research seeks to accurately ascertain the precise location of tree branches and their neighboring regions.

The evolution of object detection and image segmentation has been marked by significant advancements since the introduction of the Region-based Convolutional Neural Network (R-CNN) in 2014 \cite{girshick2014rich}. R-CNN represented a leap forward in detection accuracy by utilizing candidate regions for feature extraction and classification. Following this, the Spatial Pyramid Pooling Net (SPP-Net) \cite{he2015spatial} addressed the issue of fixed input size, allowing networks to retain more spatial information and thus improving the efficiency of the feature extraction process.

Further developments include Fast R-CNN \cite{girshick2015fast}, which enhanced both training speed and effectiveness by integrating the ROI Pooling layer, enabling feature extraction directly on the feature map. Faster R-CNN introduced the Region Proposal Network (RPN) \cite{ren2016faster}, which allowed for the generation of candidate regions and feature extraction to share computational resources, thereby significantly improving both speed and accuracy. Mask R-CNN \cite{he2017mask} added an additional branch for generating object masks, enabling pixel-level segmentation. 

For real-time object detection, the YOLO series \cite{reis2023real} \cite{wang2024yolov9} has established itself as a highly influential framework within both industrial and academic contexts, primarily due to its remarkable speed and precision. Considering its applicability to drone-based operations, we will prioritize the evaluation of the latest YOLO algorithm in our forthcoming experiments. Additionally, we will compare it with Mask R-CNN to determine the most effective solution for our needs.

\subsection{Depth Map}
Depth map generation \cite{laga2020survey} is also another crucial aspect of computer vision, enabling the inference of a scene's three-dimensional structure from one or more images. In our drone application, equipped with a stereo camera, depth maps are obtained from two distinct viewpoints. For precise pruning of branches using a pruning tool mounted on the drone, accurately identifying the tree branches and determining their distance from the drone is essential.

Depth map, representing the distance from each pixel in the image to the camera, are generated using either active or passive methods. Active methods employ sensors that emit and receive signals to measure depth, including technologies such as LiDAR, structured light\cite{wu2018squeezeseg}, and time-of-flight cameras\cite{kolb2010time}. Conversely, passive methods rely on existing optical information, utilizing techniques such as stereo matching\cite{wang2011obtaining}, multi-view geometry, and monocular depth estimation\cite{hartley2003multiple}.

Since the drone is equipped exclusively with a stereo camera, the methodology is inherently constrained to the use of one or two cameras. Consequently, stereo matching generates depth maps by deriving values through triangulation, leveraging the parallax effect between the cameras. To further elucidate this process, we will proceed with a mathematical formulation to offer a more rigorous and precise explanation of how depth maps are generated using stereo vision.\cite{hartley2003multiple}\cite{szeliski2022computer}
\begin{figure}[htbp]
\centering
\includegraphics[width=8cm]{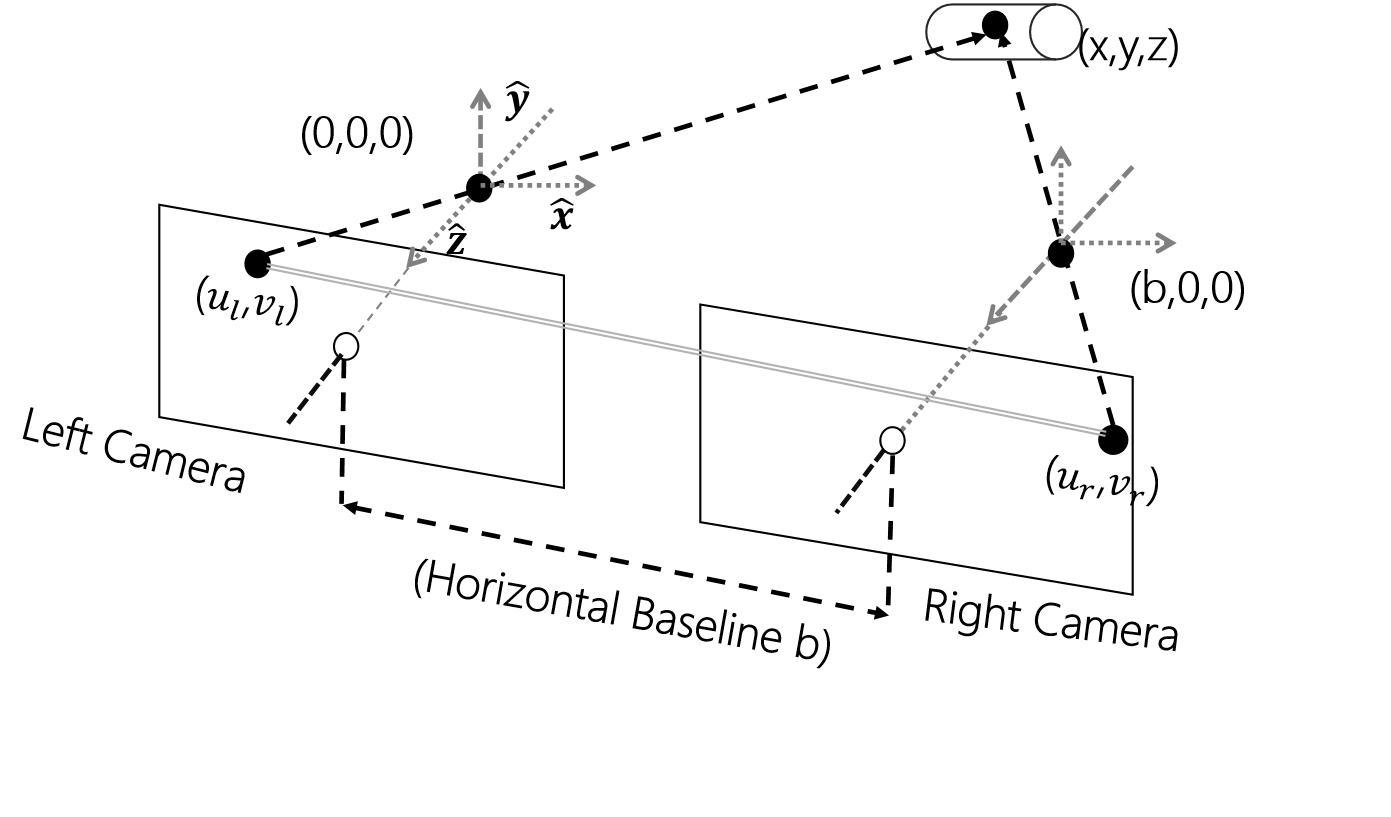}
\caption{Triangulation using Two Cameras to Obtain the Depth Map. The point \((u_l, v_l)\) represents the projection of point \(p (x, y, z)\) in three-dimensional space onto the image plane of the left camera, whereas point \((u_r, v_r)\) corresponds to the projection of the same point onto the right camera’s image plane. The variable \(b\) denotes the baseline distance separating the left and right cameras.\(\hat{x}\), \(\hat{y}\), and \(\hat{z}\) represent the three axes of the camera or world coordinate frame, corresponding to the x, y, and z directions.
}
\end{figure}
In a stereo vision system, the intrinsic camera parameters and image coordinates allow us to determine the physical properties of the scene. So we set the focal lengths of the camera along the x-axes and y-axes are denoted as $f_x$ and $f_y$, respectively. The parameters $o_x$ and $o_y$ represent the horizontal and vertical offsets of the image center from the top-left corner of the camera's image sensor. Specifically, $o_x$ refers to the horizontal displacement from the left edge of the sensor to the optical center's projection, while $o_y$ indicates the vertical displacement from the top edge of the sensor to the same point.

We then define (1) and (2) describe the position of pixel point $(u_l, v_l)$ in the left camera view, while (3) and (4) describe the position of pixel point $(u_r, v_r)$ in the right camera view.

\begin{equation}
\label{a}
u_l = f_x\frac{x}{z}+o_x
\end{equation}

\begin{equation}
\label{b}
v_l = f_y\frac{y}{z}+o_y
\end{equation}

\begin{equation}
\label{c}
u_r = f_x\frac{x-b}{z}+o_x
\end{equation}

\begin{equation}
\label{d}
v_r = f_y\frac{y}{z}+o_y
\end{equation}

\noindent Combining (1) and (2), as well as (3) and (4), we can obtain its pixel coordinates as (5).
\begin{equation}
\label{e}
\begin{aligned}
(u_l,v_l) = (f_x\frac{x}{z}+o_x, f_y\frac{y}{z}+o_y)\\
(u_r,v_r) = (f_x\frac{x-b}{z}+o_x, f_y\frac{y}{z}+o_y) 
\end{aligned}
\end{equation}
\noindent Based on the pixel coordinates of the left and right cameras, we can find the coordinates of the object in three dimensions $(x,y,z)$.
\begin{equation}
\label{f}
\begin{aligned}
x = \frac{b(u_l-o_x)}{(u_l-u_r)}\\
y = \frac{bf_x(v_l-o_y)}{f_y(u_l-u_r)}\\
z = \frac{bf_x}{(u_l-u_r)}
\end{aligned}
\end{equation}

\noindent After that, we can get the formulae for disparity value and depth value.

\begin{equation}
\label{g}
Disparity: d = u_l - u_r
\end{equation}

\begin{equation}
\label{h}
Depth: z = \frac{bf_x}{(u_l-u_r)}
\end{equation}

\noindent We set the product of the baseline $b$ and the focal length of the camera in the x-direction $f_x$ to a fixed constant $W$.
\begin{equation}
\label{i}
W = b \cdot f_x
\end{equation}
Substituting $W$ into (9), we get a more concise (10).
\begin{equation}
\label{j}
z = \frac{W}{d}
\end{equation}
Since $W$ is a fixed constant, and $z$ and $d$ are inversely proportional, the larger the disparity value, the smaller the depth value. In other words, a larger disparity value indicates that the pixel point is closer to the camera. 

Among traditional methods, Block Matching (BM) \cite{li1994new}\cite{po1996novel} and Semi-Global Block Matching (SGBM) \cite{hirschmuller2007stereo} are two dominant techniques. BM is a local search-based method that calculates depth values by finding the best match within a fixed window, making it suitable for real-time applications, though it is prone to errors in sparse texture or overlapping regions. In contrast, the SGBM method introduces a semi-global cost aggregation strategy, improving the accuracy and robustness of depth estimation by optimizing pixel points across the entire image, particularly effective in handling texture-rich scenes. 

\subsection{Performance Metrics}

In assessing the performance of the object detection and segmentation model, two critical metrics must be considered: computational efficiency (measured by running time) and accuracy. For object detection and segmentation, accuracy is evaluated using mAP\textsubscript{50--95} (Mean Average Precision at Intersection over Union thresholds ranging from 50\% to 95\%). For the depth estimation component, accuracy is measured using the Root Mean Square Error (RMSE) to quantify the depth prediction error.

Let \(AP(t)\) denote the Average Precision at a specific IoU threshold \(t\), where \(t\) represents the IoU threshold varying from 0.5 to 0.95 in increments of 0.05. The variable \(n\) denotes the total number of IoU thresholds considered, typically 10\cite{lin2014microsoft}. The formula for mAP\textsubscript{50--95} is given by:
\begin{equation}
mAP\textsubscript{50--95} = \frac{1}{n} \sum_{t=0.5}^{0.95} \text{AP}(t)
\end{equation}

Furthermore, let \(n\) as the total number of data points, where \(y_i\) denotes the actual value and \(\hat{y}_i\) represents the predicted value\cite{akaike1974new}. The formula for the RMSE is given by:
\begin{equation}
\text{RMSE} = \sqrt{\frac{1}{n} \sum_{i=1}^{n} (y_i - \hat{y}_i)^2}
\end{equation}

\section{Methods}
\begin{figure}[htbp]
\centering
\includegraphics[width=8cm]{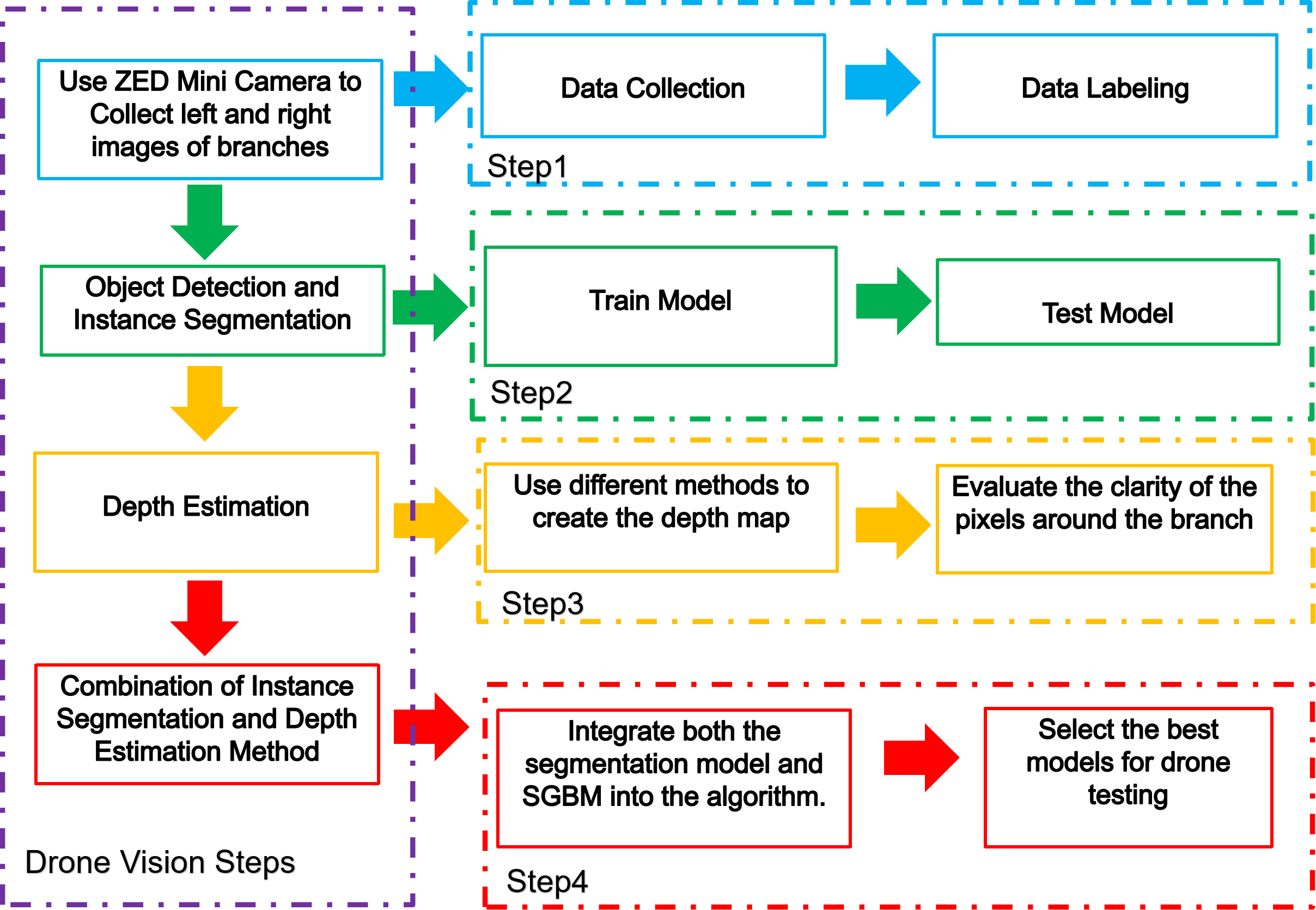}
\caption{Research flow chart}
\label{flow_chart}
\end{figure}

In this section, we systematically progress from data collection and image instance segmentation to the application of both traditional and deep learning techniques for depth map generation. By integrating these approaches, we achieve accurate detection of tree branches and estimate their distances using only a stereo vision camera. The entire workflow is illustrated in Fig.\ref{flow_chart}.

\subsection{Data Collection and Image Instance Segmentation}

In this research, we primarily collected indoor data using a ZED Mini camera\footnote{https://store.stereolabs.com/products/zed-mini}, capturing images from different corners of the laboratory at a resolution of 1920×1080. We photographed various tree branches under different lighting conditions to avoid over-idealization of the training images. So far, we have collected 61 pairs of photos (i.e., 122 images) for the training dataset and another 10 pairs for the test dataset.

After collecting the data, our process began with image labeling, specifically annotating the points around each branch. This step was essential for accurate segmentation. Given the relatively small size of the test dataset, it was critical to perform robust model testing to validate the feasibility of our approach. We initiated our experiments using Mask R-CNN, employing several backbone architectures such as ResNet-50, ResNet-101, and ResNeXt-101-32x8d. These models varied in complexity and were chosen to explore different trade-offs between speed and accuracy. The Feature Pyramid Network (FPN) and Dilated-C5 (DC5) architectures were also evaluated to assess their performance in generating masks and predicting bounding boxes.

After completing the Mask R-CNN tests, we proceeded to evaluate the dataset using YOLOv8 and YOLOv9 models of varying sizes. Once predictions were generated, we assessed the accuracy of each model using mAP\textsubscript{50--95}. In terms of computational efficiency, both Mask R-CNN and the YOLO series demonstrated impressive inference speeds, with an average processing time of approximately 10 ms per image. This performance underscores their potential for deployment in real-time applications. A comprehensive analysis of the accuracy results will be presented in the \ref{Comparison of segmentation} section.

\subsection{SGBM for Generating Depth Map}

In contrast to BM, which generates disparity maps by dividing stereo images into small blocks and performing matching along individual scan lines, SGBM optimizes the matching cost by aggregating information across multiple directions. This semi-global approach enhances accuracy and consistency by mitigating errors in textureless regions and around sharp object boundaries, resulting in a more refined and coherent disparity map. To further improve the output, Weighted Least Squares (WLS) post-processing was applied to smooth the disparity map while preserving critical edge details. The final disparity map was then converted into a depth map using Equation (10), with the corresponding results discussed in the \ref{Comparative_Depth_map} section.

\subsection{Integration of Image Instance Segmentation and Depth Map Generation}

In the earlier sections, we have thoroughly discussed both instance segmentation and depth estimation as distinct, standalone tasks. However, our ultimate objective is to enable the stereo camera mounted on the drone to simultaneously perform instance segmentation and depth estimation in order to precisely determine the spatial positions of tree branches. To achieve this, it is necessary to integrate the segmentation model and the depth map generation method.

\begin{figure}[htbp]
\centering
\subfigure[Predicted Branch Points Through YOLO]{\includegraphics[width=0.23\textwidth]{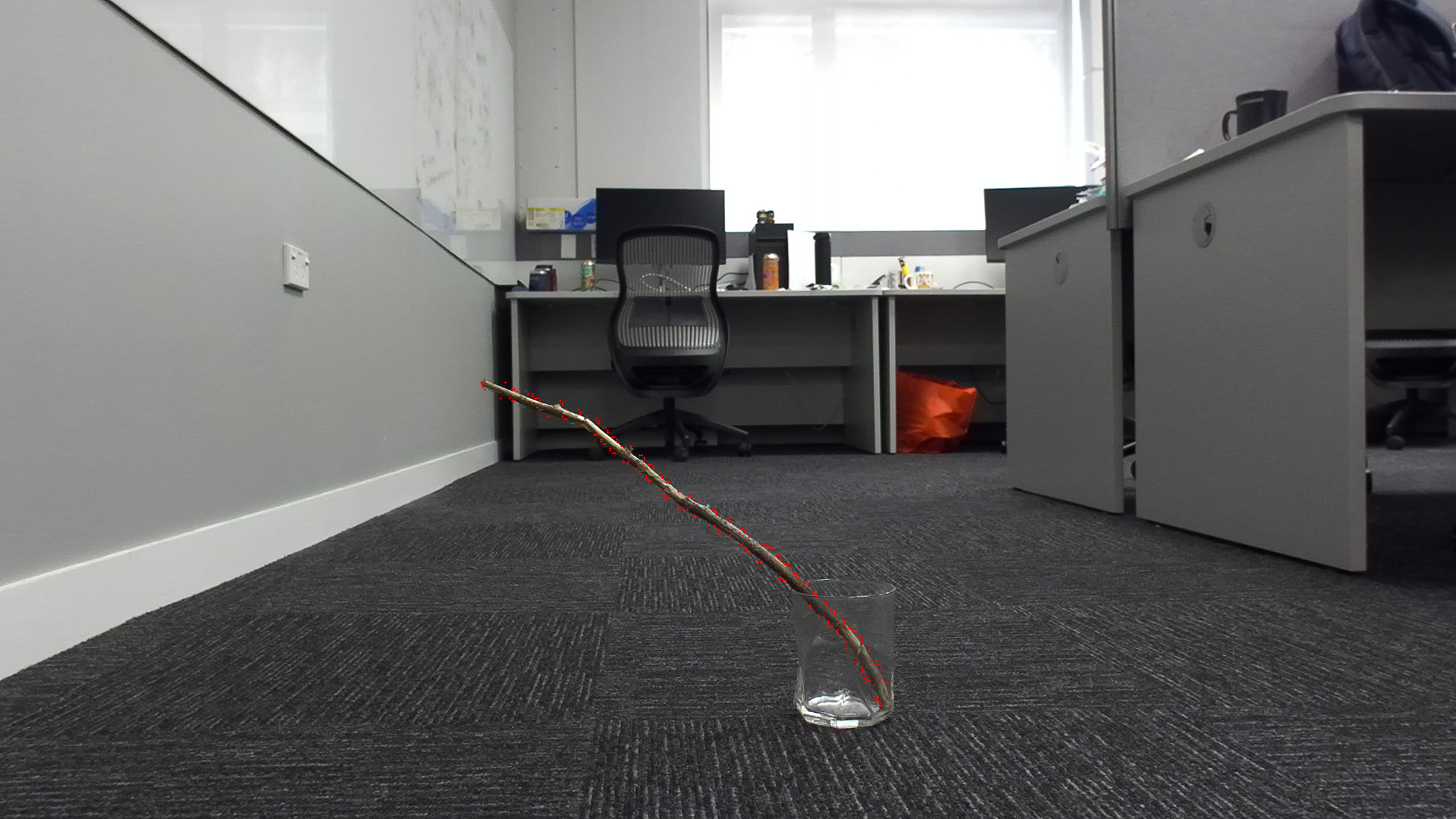}}
\subfigure[Depth Map Generated Using SGBM and WLS filter]{\includegraphics[width=0.23\textwidth]{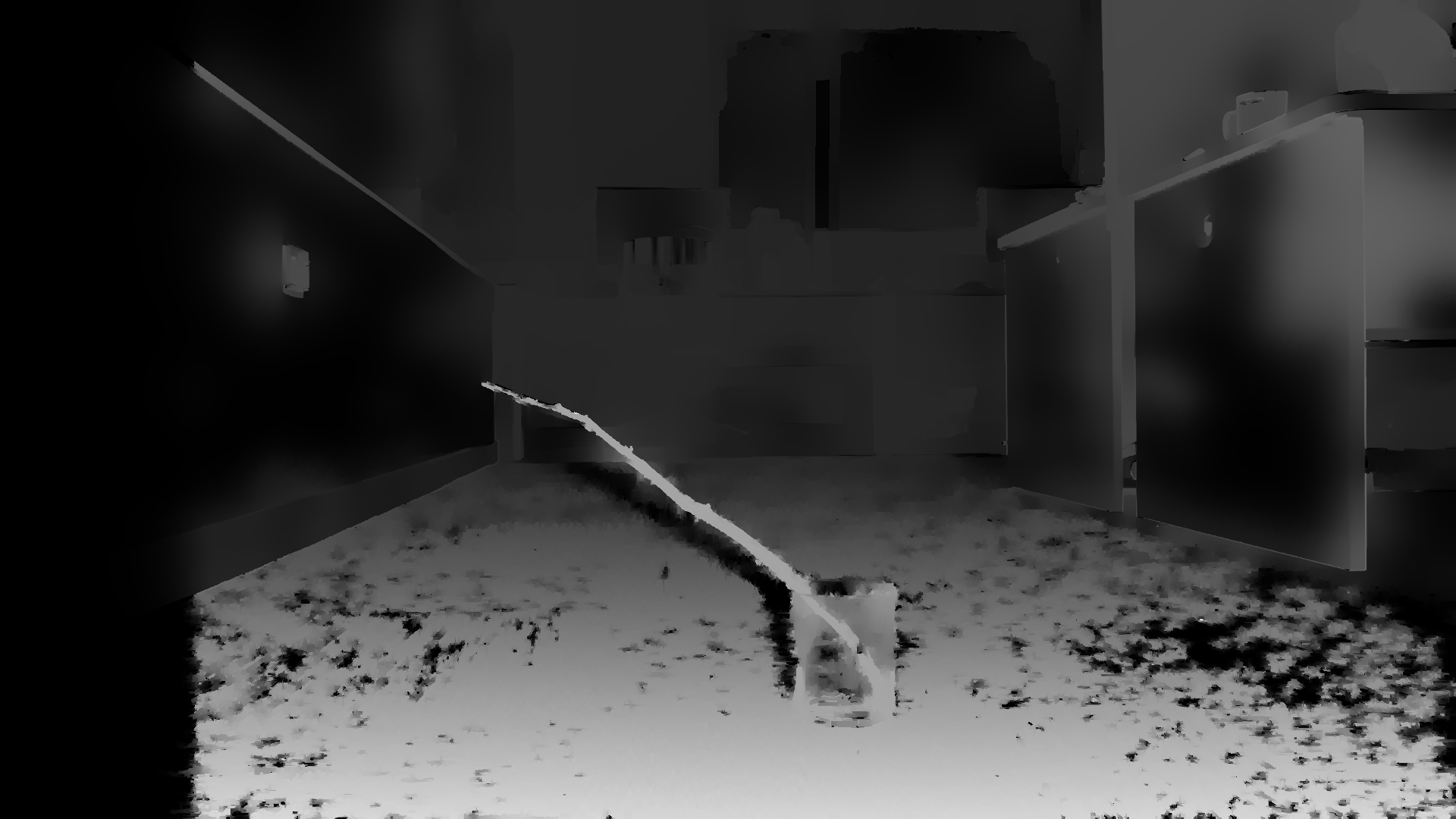}}
\caption{(a) Present the results of predicted points spaced a certain distance apart, and (b) display the depth map obtained from SGBM. Combining these allows for determining the final distance between the branches and the stereo camera.}
\label{circle_image}
\end{figure}

In Fig. \ref{circle_image}, we begin by applying a segmentation model to extract information about the points surrounding the tree branches. These points are then connected to form a continuous surface, and the coordinates of all points within this surface are mapped to their corresponding locations for depth map generation. Consequently, the depth values of all pixels on the branches are determined. Statistical analysis is then performed, and the final distance between the camera and the branches is calculated by averaging the depth values from the range where the pixel density is highest.

\section{Results and Analysis}
This section presents the results of the image segmentation and depth estimation processes, followed by an integrated analysis of their combined output. A comprehensive evaluation and interpretation of these findings are also provided, offering insights into their implications and significance within the context of the research.

\subsection{Comparative Analysis of Mask R-CNN and YOLO Models for Object Detection and Image Segmentation}
\label{Comparison of segmentation}

In the TABLE \ref{Branches}, the Mask R-CNN models utilize various backbones for feature extraction, such as ResNet-50 (R50) and ResNet-101 (R101), with options like C4 for convolutional stages, DC5 for dilated convolutions, and FPN for multi-scale feature detection. ResNeXt-101 (X101) incorporates grouped convolutions to balance accuracy and efficiency. YOLO models (v8 and v9) differ in size and computational requirements, ranging from nano (n) to extra-large (x), with segmentation capabilities indicated by "seg." The YOLOv9 models (c and e) feature further architectural enhancements for improved accuracy.

We trained for 100 epochs on our small branches dataset, yielding the results shown in TABLE \ref{Branches}. These results reveal that YOLO models significantly outperform Mask R-CNN in both box and mask mAP, with YOLOv8 and YOLOv9 achieving over 77\% mAP\textsubscript{mask50--95}, whereas Mask R-CNN struggles to reach 12\%. This highlights YOLO's superior performance in real-time detection and segmentation tasks on the branches dataset, likely due to its end-to-end design, while the two-stage approach of Mask R-CNN proves less effective for this task.

\begin{table}[h!]
\centering

\caption{Performance Comparison of Mask R-CNN and YOLO Models on our Branches Dataset(Only trained for 100 Epochs)}
\label{Branches}
\begin{tabular}{l cc} 
\toprule
\textbf{model} & \multicolumn{2}{c}{\textbf{Branches}} \\
\midrule
& mAP\textsubscript{box50--95}  & mAP\textsubscript{mask50--95} \\
\midrule
Mask R-CNN R50-C4  & 76.86 & 0.06 \\
Mask R-CNN R50-DC5  & 77.54 & 9.16 \\
Mask R-CNN R50-FPN  & 79.19 & 6.75 \\
Mask R-CNN R101-C4  & 88.05 & 0.05 \\
Mask R-CNN R101-DC5  & 79.12 & 9.94 \\
Mask R-CNN R101-FPN & 84.09 & 2.95 \\
Mask R-CNN X101-FPN & 85.52 & 11.55 \\
YOLOv8n-seg & 98.9 & 77.4 \\
YOLOv8s-seg  & 99.5 & 82.0 \\
YOLOv8m-seg  & 99.6 & 81.6\\
YOLOv8l-seg  & 99.2 & 80.1 \\
YOLOv8x-seg  & 98.7 & 77.1 \\
YOLOv9c-seg  & 98.9 & 80.9 \\
YOLOv9e-seg  & 98.8 & 80.0 \\
\bottomrule
\end{tabular}%
\end{table}

\subsection{SGBM depth map generation process}
\label{Comparative_Depth_map}

We capture the original left and right images using the ZED Mini stereo camera. These images undergo preprocessing techniques, such as smoothing, to improve their quality and reduce noise. We then apply the SGBM, followed by WLS filtering, to produce the final disparity map. The depth map is subsequently generated using equation (10). The results demonstrate that most points on the tree branches are accurately represented with clear depth information. However, some regions exhibit mismatches on the branches, indicating areas of incorrect correspondence or depth estimation.

\begin{figure}[htbp]
\centering
\subfigure[original left image]{\includegraphics[width=0.23\textwidth]{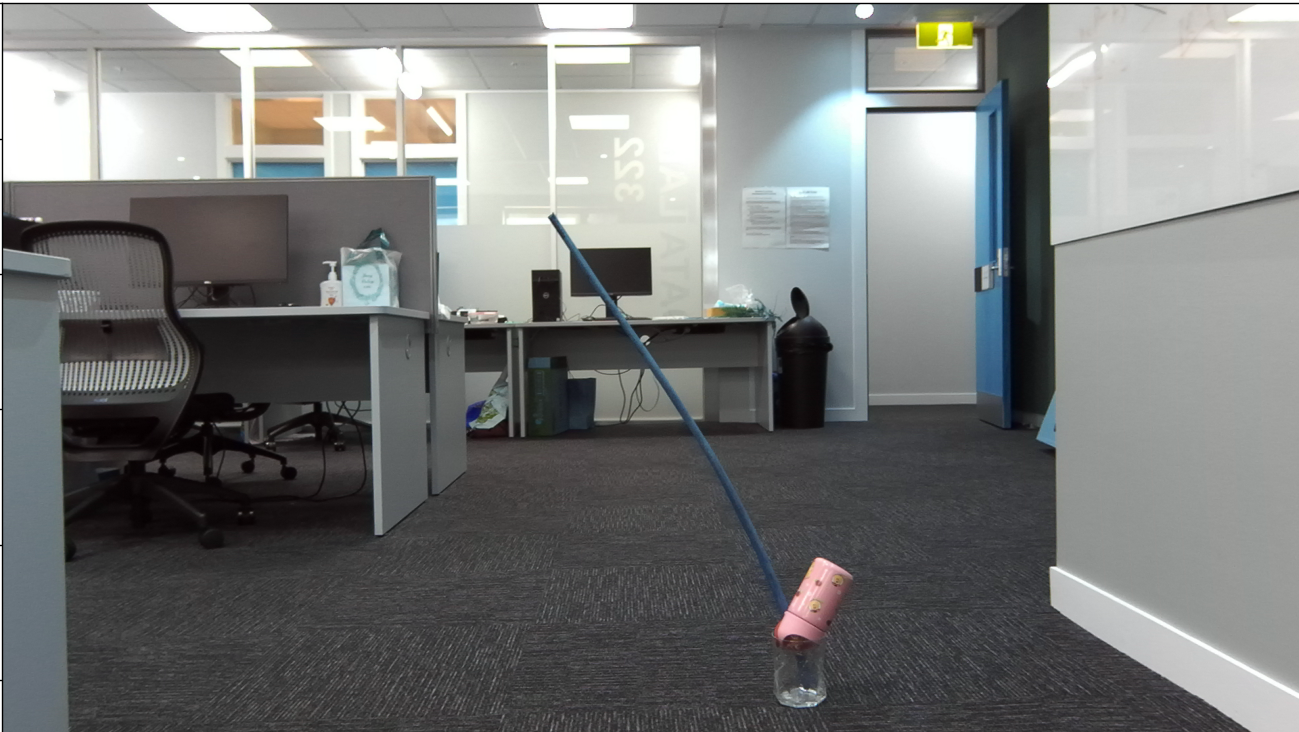}} 
\subfigure[original right image]{\includegraphics[width=0.23\textwidth]{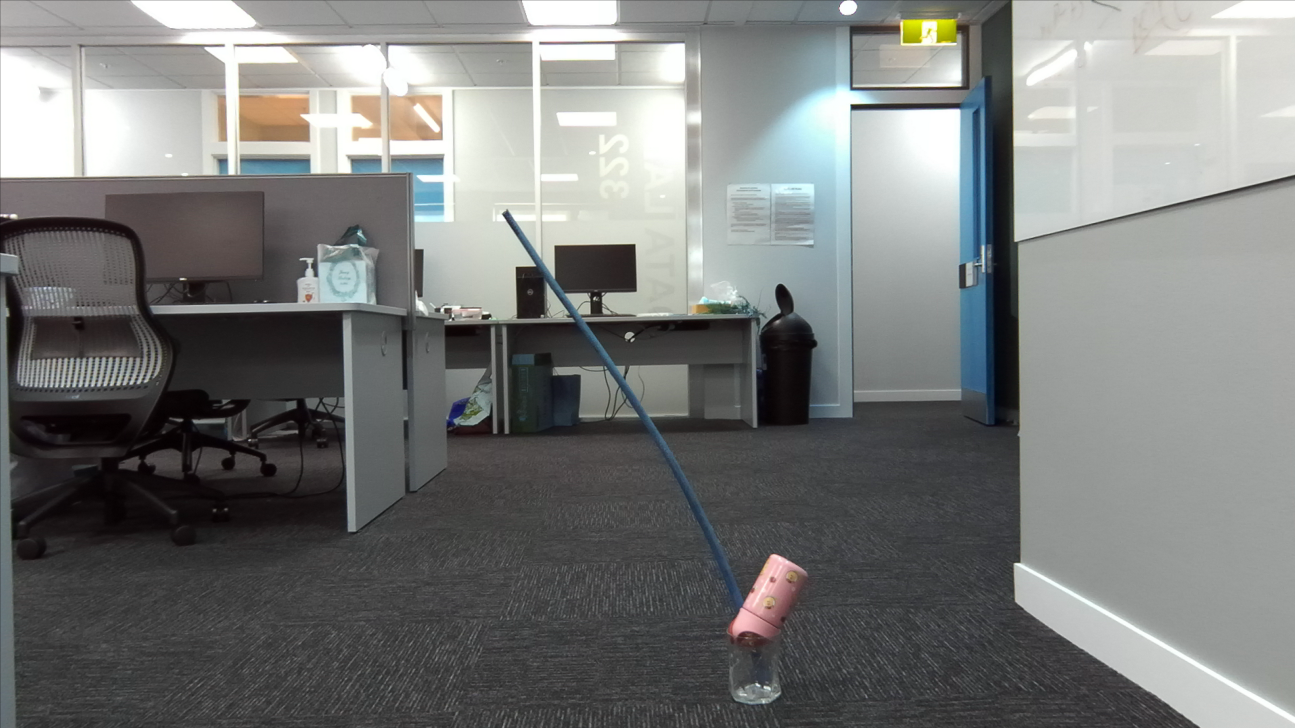}}
\subfigure[left imageafter pre-processed]{\includegraphics[width=0.23\textwidth]{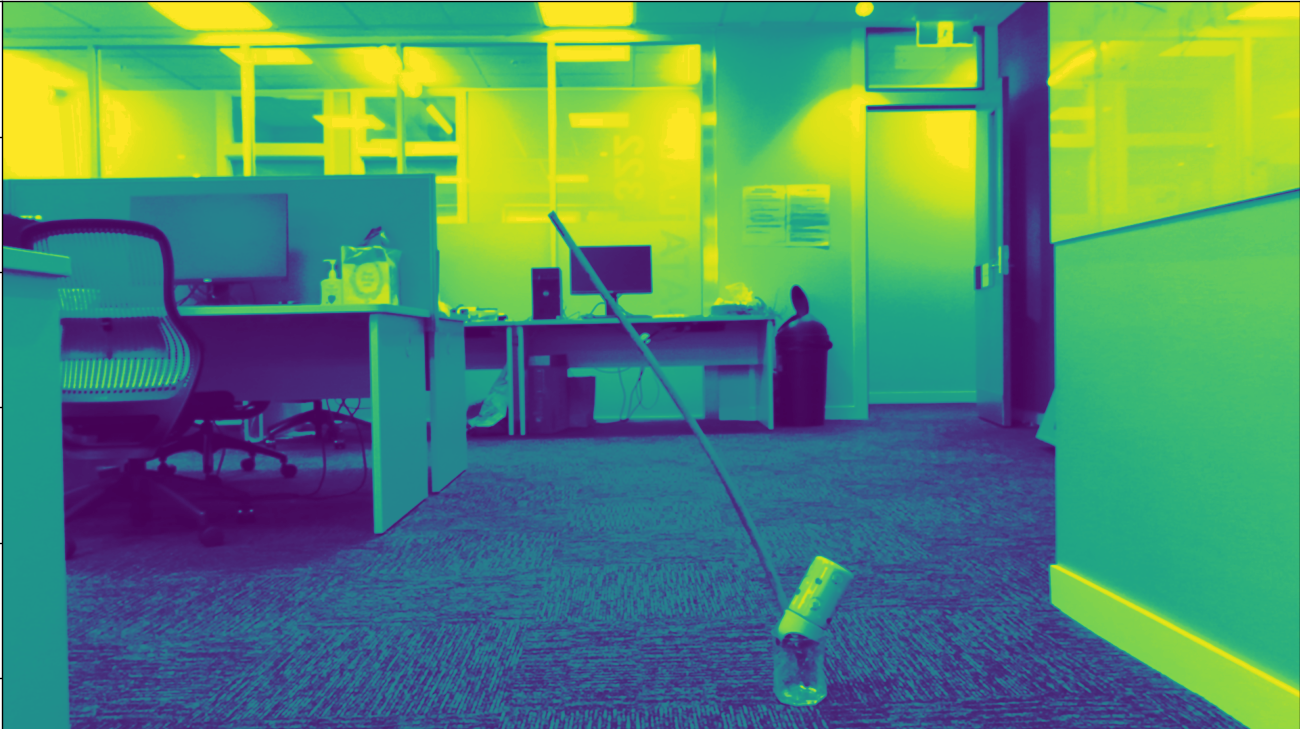}}
\subfigure[right image after pre-processed]{\includegraphics[width=0.23\textwidth]{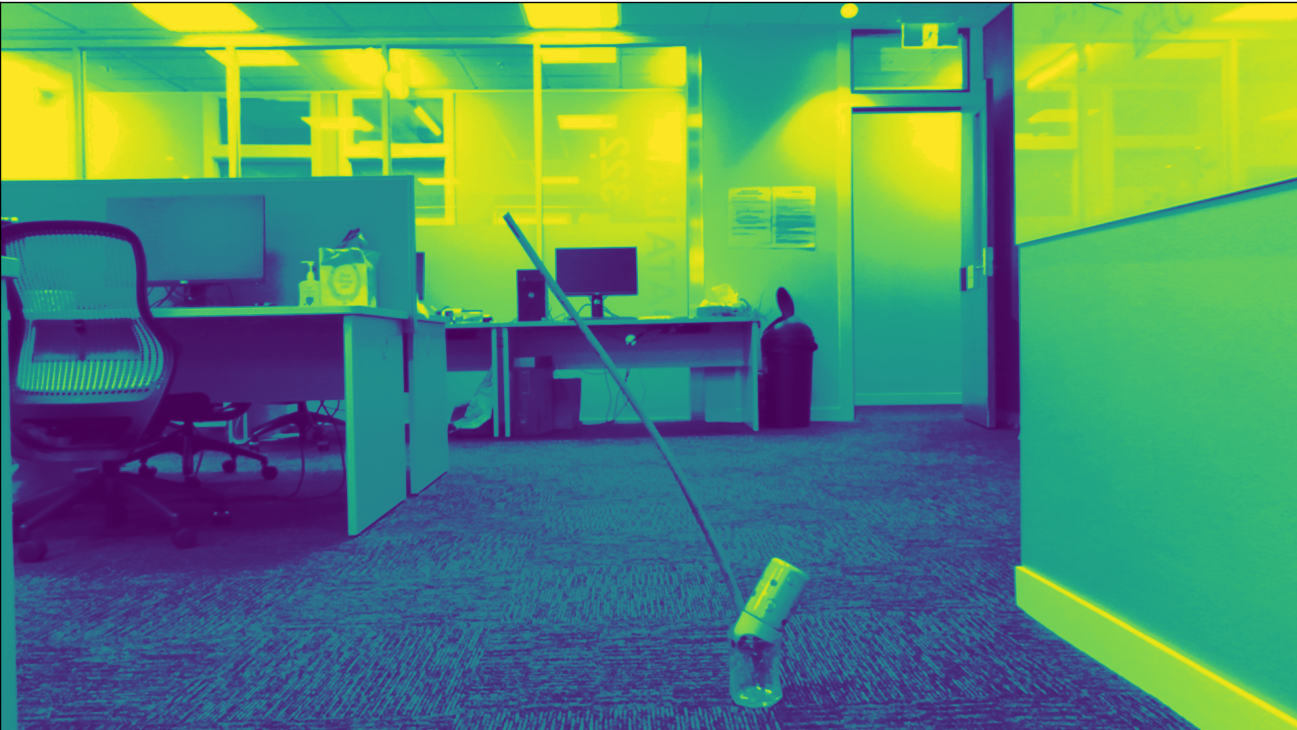}}
\subfigure[disparity map through SGBM]{\includegraphics[width=0.23\textwidth]{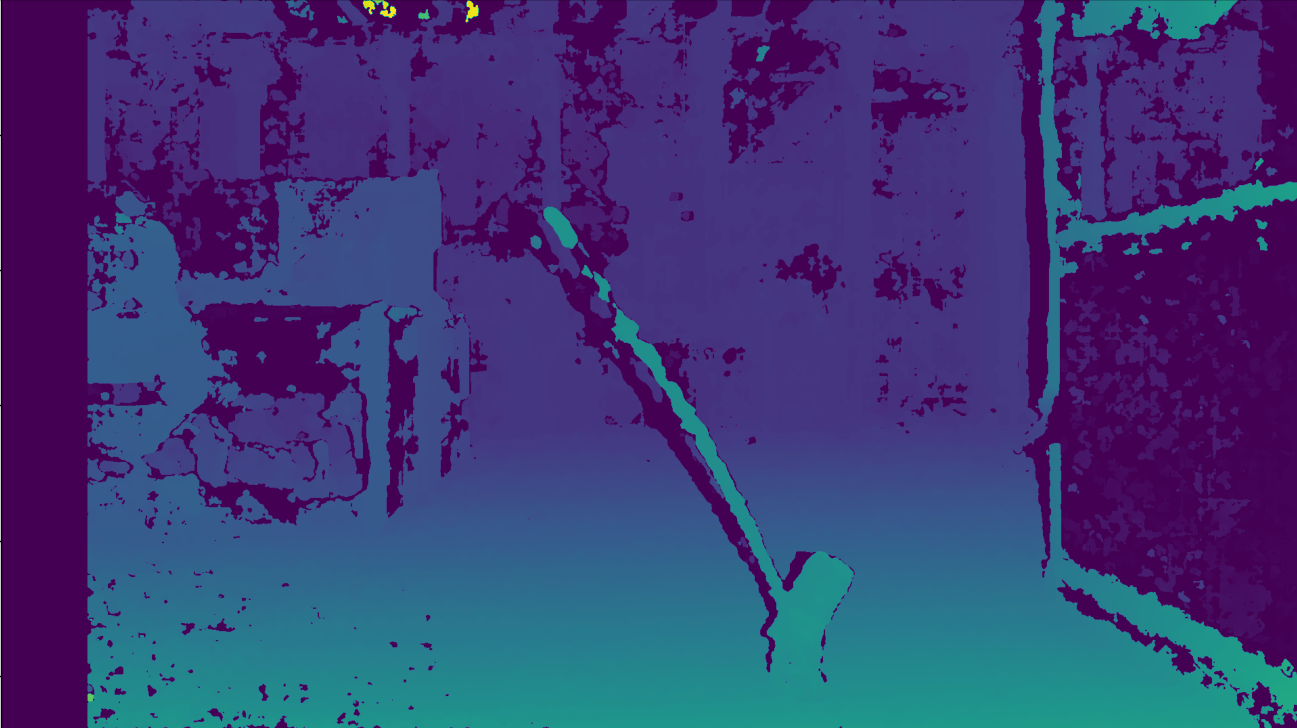}}
\subfigure[disparity map processed by Weighted Least Squares]
{\includegraphics[width=0.23\textwidth]{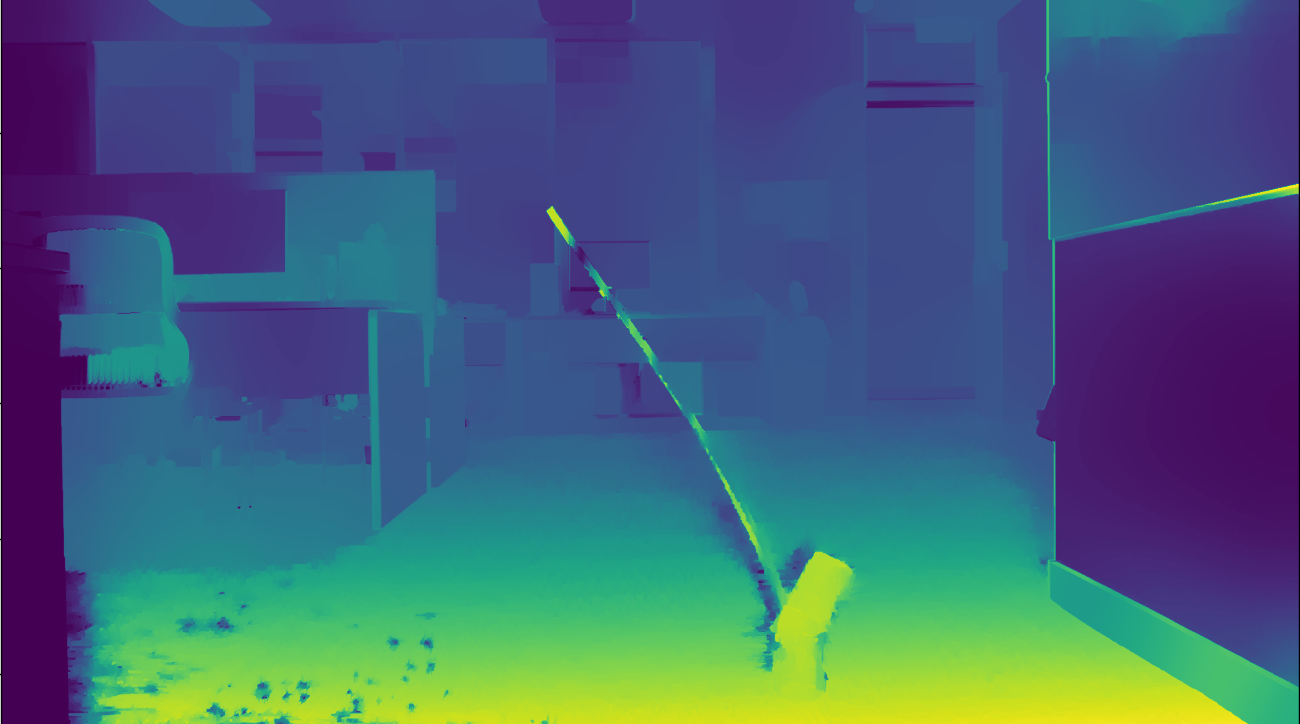}}
\caption{Show the original images, pre-processed images and disparity maps, (a) and (b) are the original left image and right image, after prepocessed we can get the (c) and (d), then we use SGBM to create the disparity map (e), then we through the WLS to get the (f).}
\label{Traditional}
\end{figure}

\subsection{Final Results Achieved by Combining YOLO with SGBM Using the First Combination Method}

\begin{figure}[htbp]
\centering
\subfigure[SGBM Generated Depth Map at 1m Distance from Branch to Camera]{\includegraphics[width=0.23\textwidth]{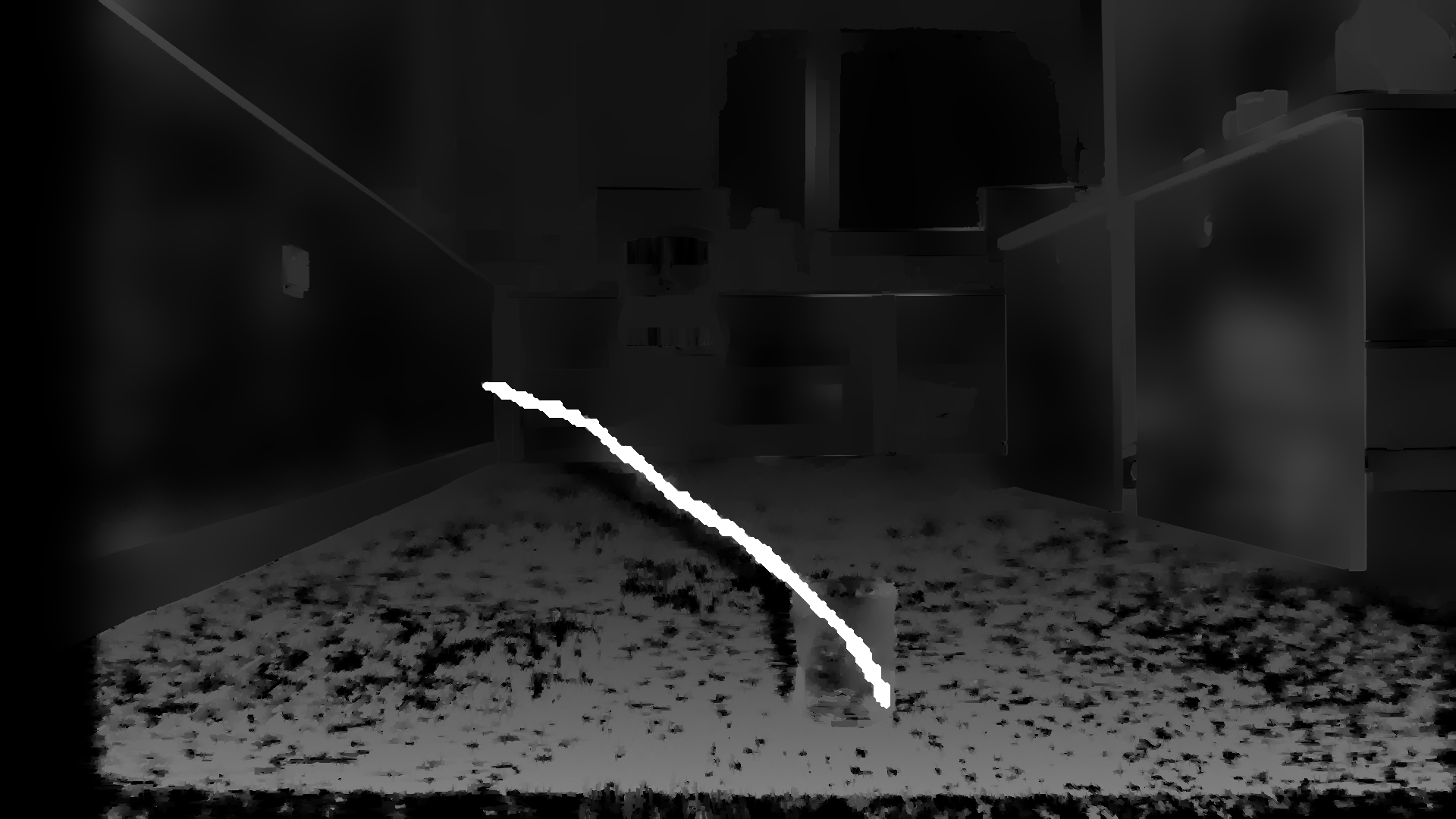}}
\subfigure[SGBM Generated Depth Map at 1.5m Distance from Branch to Camera]{\includegraphics[width=0.23\textwidth]{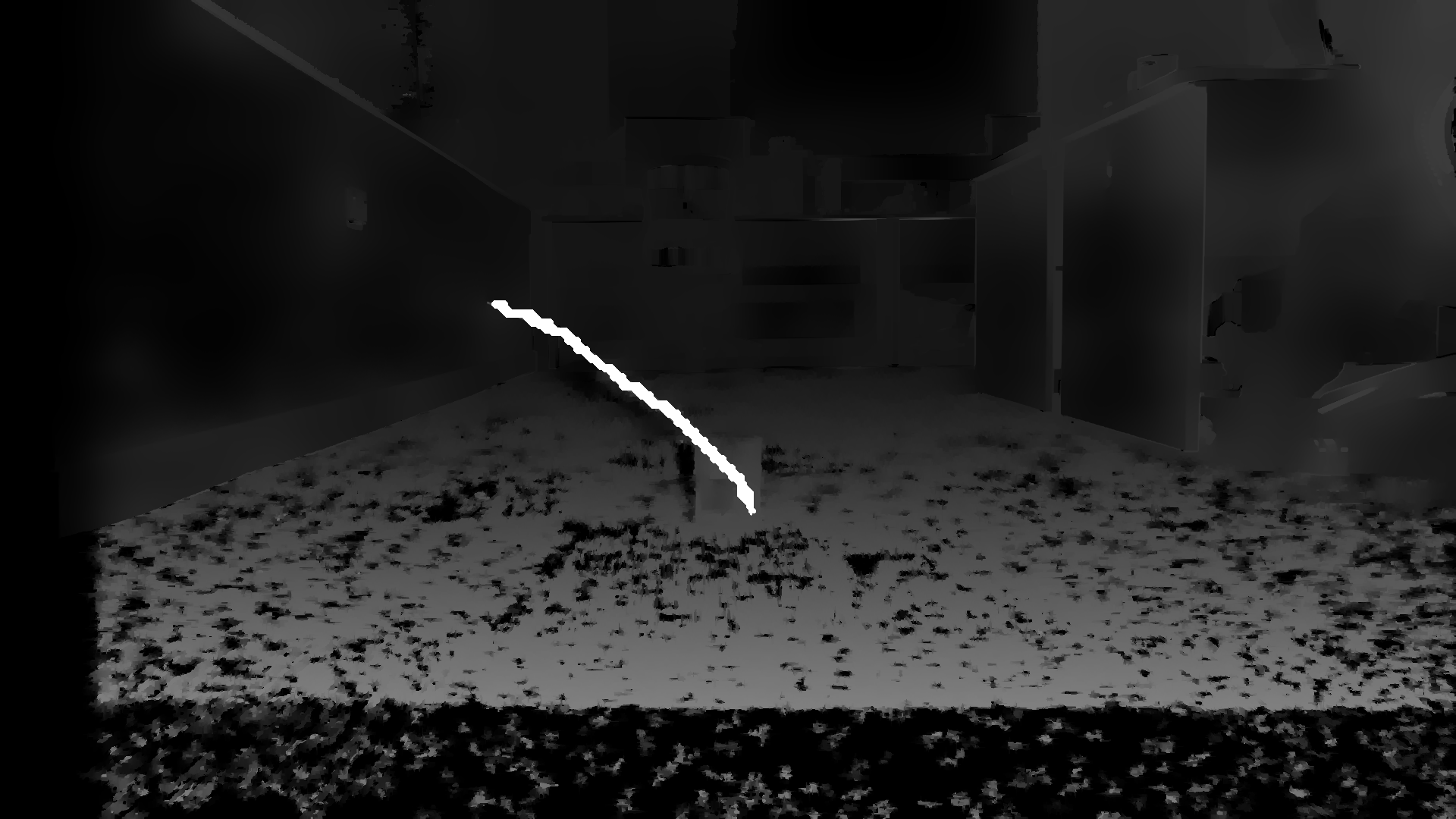}}
\subfigure[SGBM Generated Depth Map at 2m Distance from Branch to Camera]{\includegraphics[width=0.23\textwidth]{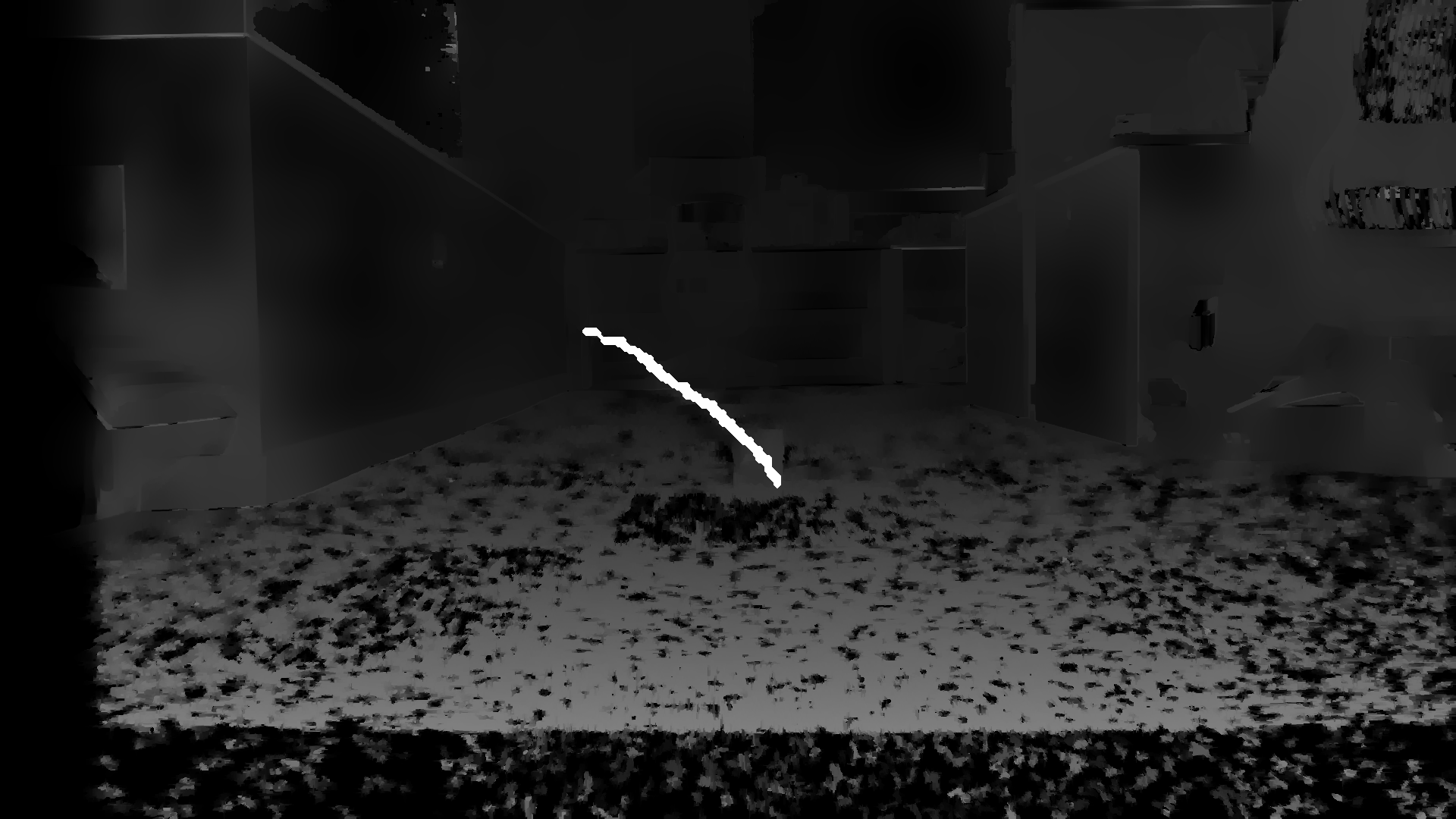}}
\subfigure[Histogram of SGBM Generated Depth Map at 1m Distance Between Branch and Camera]{\includegraphics[width=0.23\textwidth]{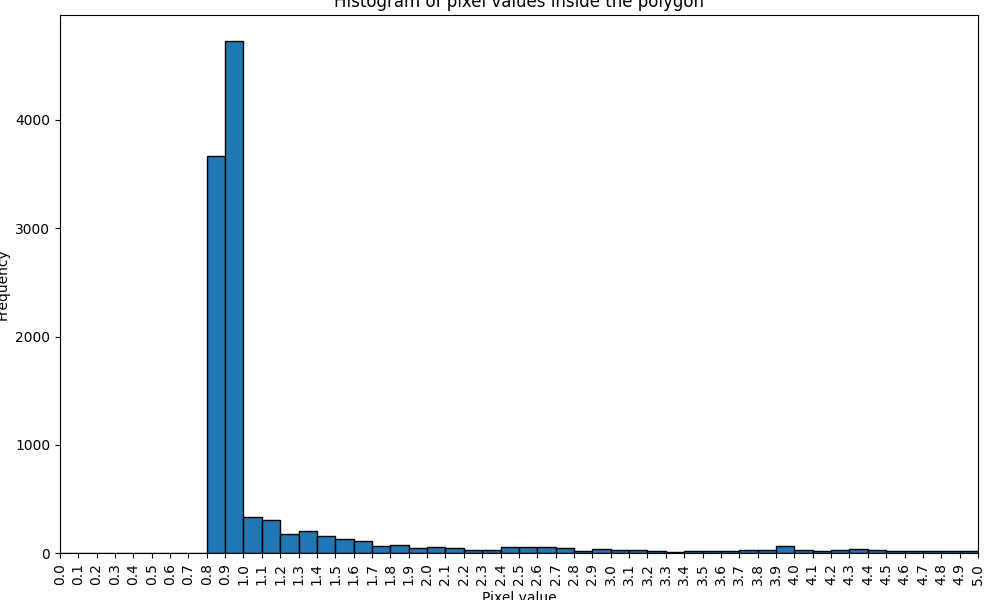}}
\subfigure[Histogram of SGBM Generated Depth Map at 1.5m Distance Between Branch and Camera]{\includegraphics[width=0.23\textwidth]{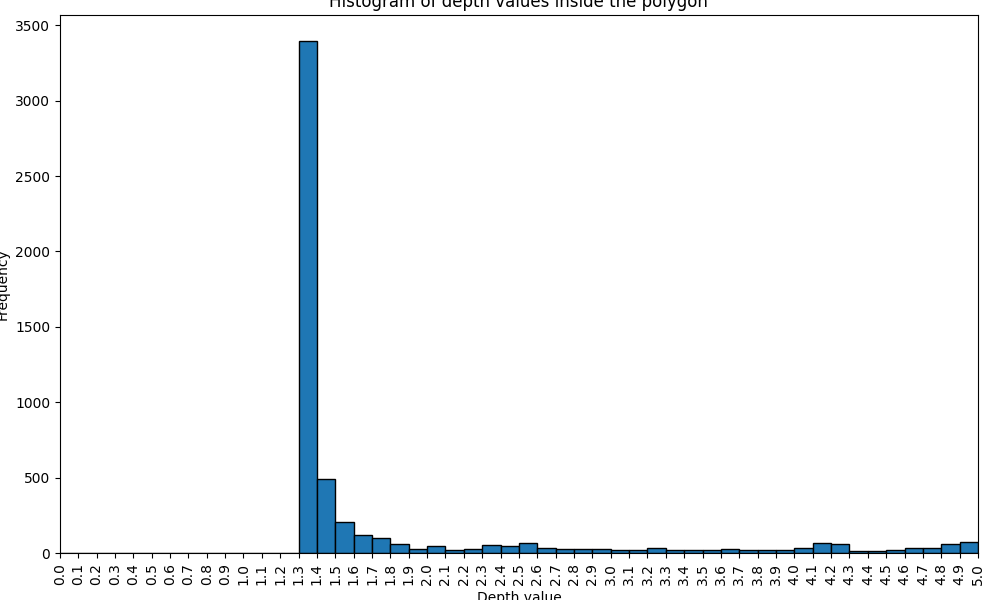}}
\subfigure[Histogram of SGBM Generated Depth Map at 2m Distance Between Branch and Camera]
{\includegraphics[width=0.23\textwidth]{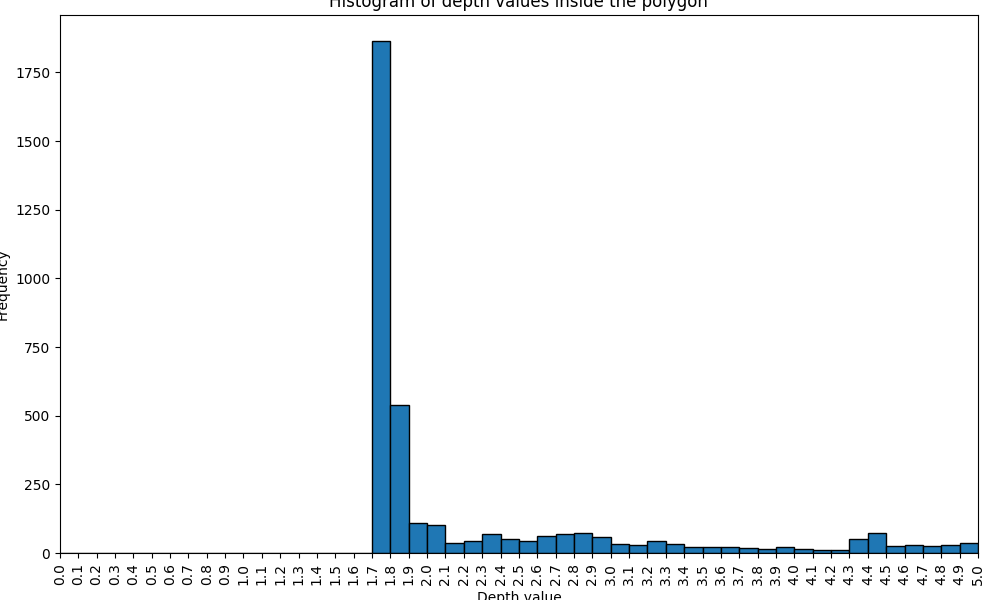}}
\caption{Comparison of YOLO Combined with SGBM at the Same Instance Across Varying Distances. with (a)-(c) Representing SGBM at 1m, 1.5m, and 2m, and (d)-(f) Representing Distribution Plots of YOLO Combined with SGBM at 1m, 1.5m, and 2m.}
\label{SGBM_and_distribution}
\end{figure}

We selected YOLO as our instance segmentation model and combined it with SGBM for depth estimation. Tests were conducted at distances of 1m, 1.5m, and 2m, and the distribution of results is illustrated in Fig. \ref{SGBM_and_distribution}. The analysis indicates that SGBM accurately detects most points within the tested distance ranges, with the majority of points correctly identified at 2 meters. Moreover, the process is completed in less than 1 second, showcasing its efficiency. Therefore, for tasks such as branch detection, using a camera-equipped drone with the YOLO and SGBM combination provides an accurate and time-efficient solution for determining branch-to-drone distances.

\section{Conclusions}

This research underscores the critical importance of computer vision techniques in accurately detecting tree branch depth information, which is essential for precision drone-assisted pruning. The research focuses on two primary components: branch detection and segmentation, and depth map generation. In the detection phase, various architectures of Mask R-CNN and YOLO were compared, with YOLO ultimately selected for its superior performance. For depth map generation, a comprehensive analysis revealed that SGBM provided satisfactory accuracy and robustness. While deep learning approaches can capture intricate features through complex neural networks, SGBM was chosen for its efficiency and reliability in our application. By integrating advanced branch detection with accurate depth maps generated by SGBM, the research enables precise measurement of distances between branches and the drone, facilitating more accurate and efficient pruning operations.

\bibliographystyle{IEEEtran} 

\vspace{12pt}

\end{document}